\documentclass{llncs}
\usepackage{url}
\usepackage{latexsym}
\usepackage[usenames, dvipsnames]{color}
\usepackage{graphicx}
\graphicspath{ {images/} }
\usepackage{booktabs, multicol, multirow,floatrow}
\usepackage{anyfontsize}
\usepackage{tabularx}
\usepackage{amsmath}
\usepackage{enumitem,tcolorbox}

\usepackage{float}
\floatstyle{plaintop}
\restylefloat{table}

\usepackage{epsf}
\usepackage{xcolor}
\usepackage{soul}

\begin{document}

\title{Detecting Sockpuppets\\ in Deceptive Opinion Spam}

\author{
Marjan Hosseinia and Arjun Mukherjee}
\institute{University of Houston\\
           Department of Computer Science\\
           Houston, TX, USA\\
           mhosseinia@uh.edu, arjun@cs.uh.edu}

\maketitle

\begin{abstract}
This paper explores the problem of sockpuppet detection in deceptive opinion spam using authorship attribution and verification approaches. Two methods are explored. The first is a feature subsampling scheme that uses the KL-Divergence on stylistic language models of an author to find discriminative features. The second is a transduction scheme, spy induction that leverages the diversity of authors in the unlabeled test set by sending a set of spies (positive samples) from training set to retrieve hidden samples in the unlabeled test set using nearest and farthest neighbors. Experiments using ground truth sockpuppet data show the effectiveness of the proposed schemes.

\end{abstract}

\section{Introduction}

Deceptive opinion spam refers to illegitimate activities, such as  writing fake reviews, giving fake ratings, etc., to mislead consumers. While the problem has been researched from both linguistic \cite{ott2011finding,feng2012syntactic} and behavioral \cite{mukherjee2013spotting,lim2010detecting} aspects, the case of sockpuppets still remains unsolved. A sockpuppet refers to a physical author  using multiple aliases (user-ids) to inflict opinion spam to avoid getting filtered. Sockpuppets are particularly difficult to detect by existing opinion spam detection methods as a sockpuppet invariably uses a user-id only a few times (often once) thereby limiting context per user-id. Deceptive sockpuppets may thus be considered as a new frontier of attacks in opinion spam.\par
 However, specific behavioral techniques such as  Internet Protocol (IP) and session logs based detection in \cite{li2015analyzing} and group spammer detection in \cite{mukherjee2012spotting} can provide important signals to probe into few ids that form a potential sockpuppet. Particularly, some strong signals such as using same  IP and session logs, abnormal keystroke similarities, etc. (all of which are almost always available to a website administrator) can render decent confidence that some reviews are written by one author masked behind a sockpuppet. This can render a form of “training data” for identifying that sockpuppeter; and the challenge is to find other fake reviews which are also written by the same author but using different aliases in future. Hence, the problem is reduced to an author verification problem. Given a few instances (reviews) written by a (known) sockpuppet author $a$, the task is to build an Author Verifier, $AV_a$ (classifier) that can determine whether another (future) review is also written by $a$ or not.
This problem is related to authorship attribution (AA)\cite{stamatatos2009survey} where the goal is to identify the author of a given document from a closed set of authors. However, having short reviews with diverse topics render traditional AA methods, that mostly rely on content features, not very effective (see section 7). While there have been works in AA for short texts such as tweets in  \cite{layton2010authorship} and with limited training data  \cite{luyckx2008authorship}, the case for sockpuppets is different because it involves deception. Further, in reality sockpuppet detection is an open set problem (i.e., it has an infinite number of classes or authors) which makes it very difficult if not impossible to have a very good representative sample of the negative set for an author. In that regard, our problem bears resemblance with authorship verification \cite{koppel2004authorship}.\\
In this work we first find that under traditional attribution setting, the precision of a verifier $AV_a$ degrades with the increase in the diversity and size of $\neg a$, where $\neg a$ refers to the negative set authors for a given verifier $AV_a$. This is detailed in section 4.1. This shows that the verifier struggles with higher false positive and cannot learn $\neg a$ well. It lays the ground for exploiting the unlabeled test set to improve the negative set in training. Next, we improve the performance by learning verification models in lower dimensions (section 5). Particularly, we employ a feature selection scheme, $\Delta$KL Parse Tree Features (henceforth abbreviated as $\Delta$KL-PTFs) that exploits the KL-Divergence of the stylistic language models (computed using PTFs) of $a$ and $\neg a$.
Lastly, we address the problem by taking advantage of transduction (section 6). The idea is to simply put a carefully selected subset of positive samples, reviews authored by $a$ (referred to as a spy set) from the training set to the unlabeled test set (i.e., the test set without seeing the true labels) and extract the nearest and farthest neighbors of the members in the spy set. These extracted neighbors (i.e., samples in the unlabeled test set which are close and far from the samples in the spy set) are potentially positive and negative samples that can improve building the verifier $AV_a$. This process is referred to as \textit{spy induction}. The basic rationale is that since all samples retain their identity, a good distance metric should find hidden positive and negative samples in the unlabeled test set. The technique is particularly effective for situations where training data is limited in size and diversity.  Although both spy induction and traditional transduction \cite{vapnik2013nature}  exploit the assumption of implicit clusters in the data \cite{chapelle2005semi}, there is a major difference between these two schemes; Spy induction focuses on sub-sampling the unlabeled test set for potential positive and negative examples to grow the training set whereas traditional transduction uses the entire unlabeled test set to find the hyperplane that splits training and test sets in the same manner \cite{joachims1999transductive}. Our results show that for the current task, spy induction significantly outperforms traditional transduction and other baselines across a variety of classifiers and even for cross domains.

\section{Related Work}
\textbf{Authorship Attribution (AA):} AA solves the attribution problem on a closed set of authors using text categorization. Supervised multi-class classification algorithms with lexical, semantic, syntactic, stylistic, and character n-gram features have been explored in \cite{graham2005segmenting,gamon2004linguistic,sapkota2015not}. In \cite{qian2016tri}, a tri-training method was proposed to solve AA under limited training data that extended co-training using three views: lexical, character and syntactic. The method however assumes that a large set of unlabeled documents authored by the same given closed set of authors are available which is different from our sockpuppet verification. In \cite{seroussi2012authorship}, latent topic features were used to improve attribution. This method also requires larger text collection per author to discover the latent topics for each author which is unavailable for a sockpuppet.\par
\textbf{Authorship Verification (AV):} In AV, given writings of an author, the task is to determine if a new document is written by that author or not. Koppel and Schler, (2004)\cite{koppel2004authorship} explored the problem on American novelists using one-class classification and ``unmasking" technique. Unmasking exploits the rate of deterioration of the accuracy of learned models as the best features are iteratively dropped. In \cite{koppel2014determining}, the task was to determine whether a pair of blogs were written by the same author. Repeated feature sub-sampling was used to determine if one document of the pair allowed selecting the other among a background set of ``imposters" reliably. Although effective unmasking requires a few hundred word texts to gain statistical robustness and was shown to be ineffective for short texts (e.g., reviews) in \cite{sanderson2006short}.\par
\textbf{Sockpuppet Detection:} Sockpuppets were studied in \cite{solorio2013case} for detecting fake identities in Wikipedia content providers using an SVM model with word and Part Of Speech (POS) features. In \cite{qian2013identifying}, a similarity space based learning method was proposed for identifying multiple userids of the same author. These methods assume reasonable context (e.g., 30 reviews per userid). These may not be realistic in opinion spamming (e.g., \cite{mukherjee2012spotting,jindal2008opinion,fusilier2015detection}) as the reviews per userid are far less and often only one, as shown in singleton opinion spamming \cite{xie2012review}.
\section{Dataset}
\cite{gokhman2012search} reports that crowdsourcing is a reasonable method for soliciting ground truths for deceptive content. Crowdsourcing has been successfully used for opinion spam generation in various previous works \cite{ott2011finding,li2014towards,li2013identifying,Banerjee2014keystroke}. In this work, our focus is to garner ground truth samples of multiple fake reviews written by one physical author (sockpuppet). To our knowledge, there is no existing dataset available for opinion spam sockpuppets. Hence, we used Amazon Mechanical Turk.\par
Participating turkers were led to a website for this experiment where responses were captured. To model a realistic scenario such as singleton opinion spamming \cite{xie2012review}, Turkers were asked to act as a sockpuppet having access to several user-ids and each user-id was to be used exactly once to write a review as if written by that alias. The core task required writing 6 positive and 6 negative deceptive reviews, each had more than 200 words, on an entity (i.e., 12 reviews per entity). Each entity belonged to one of the three domains: hotel, restaurant and product.  We selected 6 entities across each domain for this task. Each turker had to complete the core task for two entities each per domain (i.e., 24 reviews per domain). The entities and domains were spread out evenly across 17 authors (Turkers). It took us over a month to collect all samples and the mean writing time per review was about 9 minutes.\par
To ensure original content, copy and paste was disabled in the logging website. We also followed important rubrics in \cite{ott2011finding} (e.g., restricted to US Turkers, maintaining an approval rating of at least $ 90\% $) and Turkers were briefed with the domain of deception with example fake reviews (from Yelp). All responses were evaluated manually and those not meeting the requirements (e.g., overly short, incorrect target entity, unintelligible, etc.) were discarded resulting in an average of 23 reviews per Turker per domain. The data and code of this work is available at this link \footnote{https://www.dropbox.com/sh/xybjmxffmype3u2/AAA95vdkDp6z5fnTHxqjxq5Ga?dl=0} and will be released to serve as a resource for furthering research on opinion spam and sockpuppet detection. \par
Throughout the paper, for single domain experiments, we focus on the hotel domain which had the same trends to that of product and restaurant domains. However, we report results on all domains for cross domain analysis (section 7.4).
\begin{figure}[t!]
\includegraphics[scale=0.4]{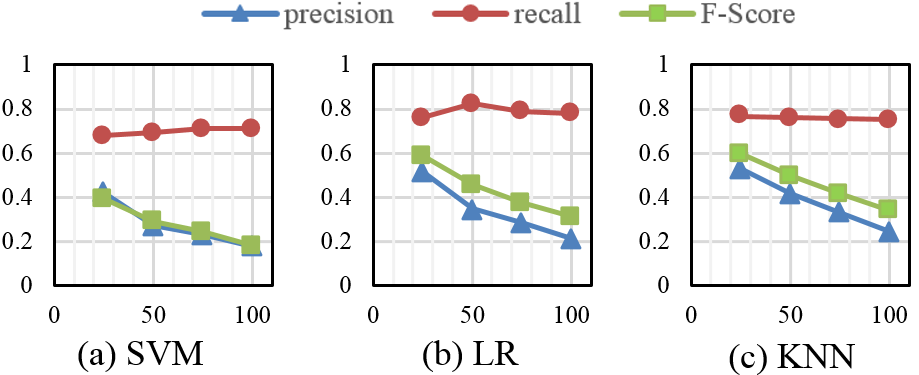}
\caption{Precision, recall and F-Score (y-axis) for different author diversity, $\lambda=25\%, 50\%, 75\%, 100\%$ (x-axis) under in-training setting.}
\end{figure}
\section{Hardness Analysis}
This section aims to understand the hardness of sockpuppet verification via two schemes.

\subsection{Employing Attribution}
An ideal verifier (classifier) for an author $a$ requires a representative sample of $\neg a$. We can approximate this by assuming a pseudo author representing $\neg a$ and populating it by randomly selecting reviews of all authors except $a$. Under the AA paradigm, this is reduced to binary classification. We build author verifiers for each author $a_i \in A=\{a_1,...,a_{17}\}$. As in AA paradigm, we use in-training setting, i.e., negative samples ($\neg a$) in both training and test sets are authored by the same closed set of 16 authors although the test and training sets are disjoint. Given our task, since there are not many documents per author to learn from, the effect of author diversity on problem hardness becomes relevant. Hence, we analyze the effect of the diversity and size of the negative set. Let $\lambda \in \{25\%,50\%,75\%,100\%\} $ be the fraction of total authors in $\neg a$ that are used in building the verifier $AV_a$. Here $\lambda$ refers to author diversity under in-training setting. We will later explore the effect of diversity under out-of-training setting (section 5). For e.g., when $\lambda=50\%$, we randomly choose 8 authors, 50\% of total 16 authors, from $\neg a$ to define the negative set for $AV_a$. Note that since we have a total of 16 authors in $\neg a$ for each $a$ and all $\lambda$ values, the class distribution is imbalanced with the negative class $\neg a$ in majority. We keep the training set balanced throughout the paper as recommended in \cite{mukherjee2013yelp} to avoid learning bias due to data skewness. We use 5-fold Cross Validation (5-fold CV) so, the training fold consists of 80\% of the positive ($a$) and equal sized negative ($\neg a$) samples. But the test fold includes the rest 20\% of positive and remaining negative samples except those in training. Under this scheme, since $\neg a$ is the majority class in the test set, accuracy is not an effective metric. For each $AV_a$, we first compute the precision, recall and F-Score (on the positive class $a$) using 5-fold CV. Next, we average the results across all authors using their individual verifiers (Figure 1). This scheme yields us a robust measure of performance of sockpuppet verification across all authors and is used throughout the paper.\par  We report results of Support Vector Mechine (SVM), Logistic Regression (LR) and k-Nearest Neighbor (kNN) classifiers (using the libraries LIBSVM \cite{CC01a} for SVM with RBF kernel, LIBLINEAR\cite{REF08a} for LR with L2 regularization and WEKA \footnote{http://www.cs.waikato.ac.nz/ml/weka/} for kNN with k=3  whose parameters were learned via CV). The feature space consists of lexical units (word unigram) and Parse Tree Features (PTF) extracted using Stanford parser \cite{klein2003accurate}  with normalized term frequency for feature value assignment.  Unless otherwise stated we use this feature set as well as the classifires setting for all experiments in this paper. We followed some rules from \cite{feng2012characterizing} in computing PTFs. The rules are generated by traversing a parse tree in three ways i) a parent node to the combination of all its non-leaf nodes, ii) an internal node to its grandparent, iii) a parent to its internal child. We also add all interior nodes to the feature space (Table 1). From Figure 1, we note:
\begin{table}[t!]
 \caption{Parse Tree Feature (PTF) Types}
 \label{tab:table1}
  \setlength{\tabcolsep}{0.15em}
  \centering
    \begin{tabular}{lll}
   
    \toprule
    \multicolumn{3}{c}{Parse tree for: ``The staff were friendly."} \\
    \midrule
    \multirow{4}[8]{*}{{\includegraphics[width=3cm,height=4cm,keepaspectratio]{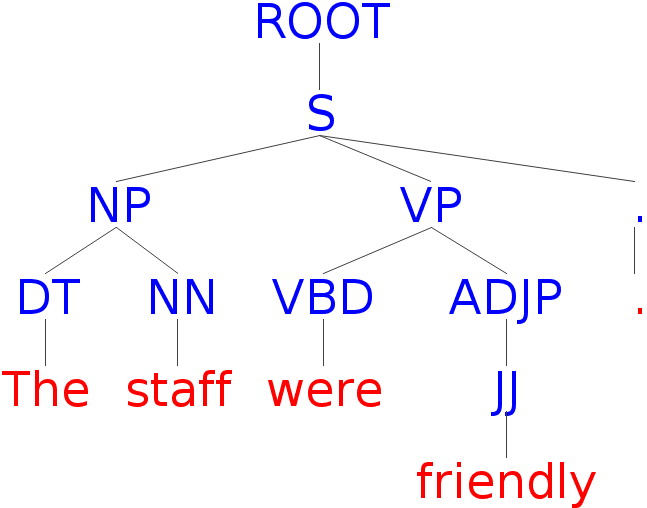}}}\\ & PTF(I) & S $\rightarrow$ NP VP \\
          & PTF(II) & JJ \^ \,ADJP $\rightarrow$ VP\\
          & PTF (III) & S$\rightarrow$ NP \\
          & Interior nodes & DT, NP \\
  
    \end{tabular}%

\end{table}%
\begin{itemize}[leftmargin=*]
\item 	With increase in diversity of negative samples, $\lambda$ of $\neg a$, the test set size and variety also increase and we find significant drops in precision across all classifiers. This shows a significant rise in false positives. In other words, as the approximated negative set approaches the universal negative set ($\widetilde{\neg a}\rightarrow\neg a$ with increase in diversity of $\neg a$), learning $\neg a$ becomes harder.
\item 	Recall, however, does not experience major changes with
increase in the diversity of negative set as it is concerned with retrieving the positive class ($a$).
\item F-Score being the harmonic mean of precision and recall, aligns with the precision performance order. We also note that F-Score in SVM and LR behave similarly followed by kNN.
\end{itemize} 
Thus, sockpuppet verification is non-trivial and the hardness increases with the increase in $\neg a$ diversity.
\subsection{Employing Accuracy and F1 on Balanced Class Distribution}
Under binary text classification and balanced class distribution, if accuracy or F1 are high, it shows that the two classes are well separated. This scheme was used in \cite{koppel2004authorship} for authorship verification. In our case, we adapt the method as follows. We consider two kinds of balanced data scenarios for a verifier for author $a$, $AV_a$: $S_1$ and $S_2$. Under $S_1$, we have the positive class $P$ that consists of half of all reviews authored by $a$ $R_a$, i.e., $P=\{r_i \in R_a;|P|=1/2 |R_a |\}$. The negative class $N_{S_1}$ comprises of the other half,  $N_{S_1}=\{r_i \in R_a-P;|N_{S_1} |=|P|\}$ and $S_1=P\cup N_{S_1}$. Under $S_2$, we keep $P$ intact but use a random sampling of $\neg a$ for its negative class, $N_{S_2}=\{r_i \in R_{\neg a};|N_{S_2} |=|P|\}$ yielding us $S_2=P\cup N_{S_2}$. Essentially, with this scheme, we wish to understand the effect of negative training set when varied from false negative $(N_{S_1})$ to approximated true negative $(N_{S_2})$. Using lexical and parse tree features and 5-fold CV we report performance under each scenario $S_1$ and $S_2$ in Table 2. We note the following:
\begin{itemize}[leftmargin=*]
\item The precision, recall, F1 and accuracy of all models under $S_2$ is higher than $S_1$. While this is intuitive, it shows for deceptive sockpuppets, writings of an author ($P$) bear separation from other sockpuppeters ($N_{S_2}$). 
\item Sockpuppet verification is a difficult problem because under balanced binary classification ($S_2$), there is just 5-10\% gain in accuracy than random (50\% accuracy). Yet it does show the models are learning some linguistic knowledge that separate $a$ and $\neg a$ and using writings of authors other than $a$ is a reasonable approximation for universal $\neg a$.  

\end{itemize}
\begin{table}[t!]
 \caption{Classification results P: Precision, R: Recall, Acc: Accuracy, F1: F-Score under two balanced data scenarios $S_1$ and $S_2$ for different classifiers.}

  \centering
  \scalebox{1}{
    \begin{tabular}{lrrrrrrrr}
    \toprule
    \multirow{2}[4]{*}{Model} & \multicolumn{4}{c|}{$S_1$}          & \multicolumn{4}{c}{$S_2$} \\
   
          & \multicolumn{1}{c}{P} & \multicolumn{1}{c}{R} & \multicolumn{1}{c}{Acc} & \multicolumn{1}{c|}{F1} & \multicolumn{1}{c}{P} & \multicolumn{1}{c}{R} & \multicolumn{1}{c}{Acc} & \multicolumn{1}{c}{F1} \\
           \midrule
    SVM   & \multicolumn{1}{c}{47.1} & \multicolumn{1}{c}{48.4} & \multicolumn{1}{c}{49.0} & \multicolumn{1}{c}{45.6} & \multicolumn{1}{c}{62.5} & \multicolumn{1}{c}{66.5} & \multicolumn{1}{c}{61.8} & \multicolumn{1}{c}{61.1} \\
    LR    & \multicolumn{1}{c}{47.4} & \multicolumn{1}{c}{46.4} & \multicolumn{1}{c}{49.5} & \multicolumn{1}{c}{44.6} & \multicolumn{1}{c}{63.5} & \multicolumn{1}{c}{67.4} & \multicolumn{1}{c}{61.7} & \multicolumn{1}{c}{62.1} \\
    kNN   & \multicolumn{1}{c}{41.9} & \multicolumn{1}{c}{57.4} & \multicolumn{1}{c}{49.8} & \multicolumn{1}{c}{44.9} & \multicolumn{1}{c}{51.0} & \multicolumn{1}{c}{68.9} & \multicolumn{1}{c}{56.1} & \multicolumn{1}{c}{53.8} \\
    \bottomrule
    \end{tabular}%
    }
   
  \label{tab:addlabel}%
\end{table}%
\section{Learning in Lower Dimensions}
From the previous experiment, it hints that in the case of deceptive sockpuppets, only a small set of features differentiate $a$ and $\neg a$. As explored in \cite{feng2012characterizing}, there often exists discriminative author specific stylistic elements that can characterize an author. However, the gamut of all PTFs per author (greater than 2000 features in our data) may be overlapping across authors (e.g., due to native language styles). To mine those discriminative PTFs, we need a feature selection scheme. We build on the idea of linguistic KL-Divergence in \cite{mukherjee2013yelp} and model stylistic elements to capture \textit{how} things are said as opposed to \textit{what} is said. The key idea is to construct the stylistic language model for author, $a$ and its pseudo author $\neg a$. Let $A$ and $\neg A$ denote the stylistic language models for author $a$ and $\neg a$ comprising the positive and negative class of $AV_a$ respectively, where $A(t)$ and $\neg A(t)$ denote the probability of the PTF, $t$ in the reviews of $a$ and $\neg a$. $KL(A||\neg A)=\sum_t (A(t)  \log_2{(A(t)/\neg A(t))})$ provides a quantitative measure of stylistic difference between $a$ and $\neg a$. Based on its definition, PTF $t$ that appears in $A$ with higher probability than in $\neg A$, contributes most to $KL(A||\neg A)$. Being asymmetric, it also follows that PTF $t^\prime $ that appears in $\neg A$ more than in $A$ contributes most to $KL(\neg A||A)$. Clearly, both of these types of PTF are useful for building $AV_a$. They can be combined by computing the per feature, $f$, $\Delta KL^f$ as follows:
\begin{align}
\Delta KL_t^f=KL_t(A_t||\neg A_t)-KL_t(\neg A_t||A_t)\\
KL_t(A_t||\neg A_t)=A(t)  \log_2{(A(t)/\neg A(t))}\\
KL_t(\neg A_t|| A_t)=\neg A(t)  \log_2{(\neg A(t)/A(t))}
\end{align}

Discriminative features are found by simply selecting the top PTF $t$ based on the descending order of $|\Delta KL_t^f |$ until $|\Delta KL_t^f |<0.01 $. This is a form of sub-sampling the original PTF space and lowers the feature dimensionality. Intuitively, as $KL_t$ is proportional to the relative difference between the probability of PTF $t$ in positive ($a$) and negative ($\neg a$) classes, the above selection scheme provides us those PTF $t$ that contribute most to the linguistic divergence between stylistic language models of $a$ and $\neg a$. \par
To evaluate the effect of learning in lower dimensions, we consider a more realistic ``out-of-training" setting instead of the in-training setting as in previous experiments. Under out-of-training setting, the classifier cannot see the writings of those authors that it may encounter in the test set. In other words test and training sets of a verifier $AV_a$ are completely disjoint with respect to $\neg a$ which is realistic and also more difficult than in-training setting. Further, we explore the effect of author diversity under out-of-training setting, $\delta$ for the negative set (not to be confused with $\lambda$ as in section $4$). For each experiment, the reviews from $\delta \%$ of all authors except the intended author, $\neg a$ participate in the training of a verifier $AV_a$ while the rest ($100-\delta \%$) authors make the negative test set. We also consider standard lexical units (word unigram) (L), L + PTF, and top $k=20\%$ (tuned via CV) PTF selected using $\chi ^2$ metric (L + PTF $\chi ^2$)  as baselines. We examine different values of $\delta \in \{25\%,50\%,75\%\}$ but not $\delta=100\%$ as that leaves no test samples due to out-of-training setting. From Table 3, we note:
\begin{itemize}[leftmargin=*]
\item For each feature space, as the $\neg a$ diversity ($\delta$) increases, across each classifier, we find gains in precision with reasonably lesser drops in recall resulting in overall higher F1. This shows that with increase in diversity in training, the verifiers reduced false positives improving their confidence. Note that verification gets harder for smaller $\delta$ as the size and skewness of the test set increases. This trend is different from what we saw in Figure 1 with $\lambda$ which referred to diversity under in-training setting.
\item Average F1 based on three classifiers (column AVG, Table 3)  improves for $\delta=25\%,50\%$ using L+PTF than L showing parse tree feature can capture style. However feature selection using $\chi ^2$ (L+PTF $\chi ^2$) is not doing well as for all $ \delta$ values there is reduction in F1 for SVM and LR. L+$\Delta KL$ PTF feature selection performs best in AVG F1 across different classifiers. It recovers the loss of PTF $\chi ^2$ and also improves over the L+PTF space by about 2-3$\%$.
\end{itemize}
\begin{table}[t!]
  \centering
 
 \scalebox{1}{
    \begin{tabular}{lcccccccccr}
    \toprule
    \multirow{2}[3]{*}{} & \multicolumn{10}{c}{\bf{ $\delta$=25\%}} \\
    
          & \multicolumn{3}{c}{\bf{SVM}} & \multicolumn{3}{c}{\bf{LR}} & \multicolumn{3}{c}{\bf{kNN}} & \multicolumn{1}{c}{\bf{AVG}} \\
          
    Feature Set & P     & R     & F1    & P     & R     & F1    & P     & R     & F1    & \multicolumn{1}{c}{F1} \\\midrule
    L     & 23.6  & 82.0    & 34.3  & 23.1  & 74.7  & 30.8  & 19.4  & 84.6  & 25.8  & \multicolumn{1}{c}{30.3} \\
    L+PTF & 25.6  & 73.4  & 35.2  & 22.9  & 82.5  & 33.4  & 24.8  & 66.7  & 24.5  & \multicolumn{1}{c}{31.0} \\
    L+PTF$\chi^2$   & 21.7  & 73.5  & 30.8  & 14.8  & 53.5  & 21.3  & 22.6  & 75.3  & 25.9  & \multicolumn{1}{c}{26.0} \\
    L+$\Delta KL$ PTF  & 25.6  & 79.2  & 36.3  & 21.7  & 80.2  & 32.1  & 22.3  & 81.5  & 27.8  & \multicolumn{1}{c}{\textbf{32.1}} \\
    \multicolumn{11}{c}{(a)} \\
    \midrule
    \multirow{2}[4]{*}{} & \multicolumn{10}{c}{\bf{$\delta$ =50\%}} \\
          & \multicolumn{3}{c}{\bf{SVM}} & \multicolumn{3}{c}{\bf{LR}} & \multicolumn{3}{c}{\bf{kNN}} & \multicolumn{1}{c}{\bf{AVG}} \\
    Feature Set & P     & R     & F1    & P     & R     & F1    & P     & R     & F1    & \multicolumn{1}{c}{F1} \\
    \midrule
    L     & 30.7  & 83.6  & 41.8  & 28.7  & 83.1  & 38.7  & 21.1  & 85.1  & 27.1  & \multicolumn{1}{c}{35.9} \\
    L+PTF & 33.2  & 73.4  & 42.7  & 30.6  & 78.1  & 40.9  & 28.0    & 73.8  & 28.8  & 37.5 \\
     L+PTF$\chi^2$   & 24.8  & 69.2  & 33.7  & 21.0    & 47.8  & 26.9  & 23.4  & 81.6  & 30.2  & 30.3 \\
     L+$\Delta KL$ PTF  & 33.7  & 75.9  & 42.8  & 31.1  & 79.4  & 41.9  & 26.9  & 79.5  & 30.3  & \textbf{38.3} \\
    \multicolumn{11}{c}{(b)} \\
    \midrule
    \multirow{2}[4]{*}{} & \multicolumn{10}{c}{\bf{$\delta$ =75\%}} \\
    
          & \multicolumn{3}{c}{\bf{SVM}} & \multicolumn{3}{c}{\bf{LR}} & \multicolumn{3}{c}{\bf{kNN}} & \multicolumn{1}{c}{\bf{AVG}} \\
    Feature Set & P     & R     & F1    & P     & R     & F1    & P     & R     & F1    & \multicolumn{1}{c}{F1} \\
    \midrule
    L     & 47.1  & 77.7  & 55.1  & 44.4  & 80.4  & 52.7  & 28.7  & 83.5  & 37.8  & \multicolumn{1}{c}{48.5} \\
    L+PTF & 51.4  & 72.6  & 56.0    & 43.7  & 78.8  & 53.0    & 28.1  & 64.8  & 31.7  & 46.9 \\
     L+PTF$\chi^2$   & 42.4  & 71.2  & 49.5  & 33.9  & 49.6  & 36.4  & 35.6  & 79.8  & 40.0    & 42.0 \\
      L+$\Delta KL$ PTF  & 50.5  & 71.9  & 56.2  & 46.3  & 79.4  & 54.9  & 42.2  & 80.8  & 46.1  & \textbf{52.4} \\
       \multicolumn{11}{c}{(c)} \\
    \bottomrule
    
    \end{tabular}%
    }
    \caption{P: Precision, R: Recall, F1: F-Score for out-of-training with different values of $\delta$ for three classifiers. AVG reports the average F1 across three classifiers. Feature Set: L: Lexical unit (word unigram), PTF: Parse tree feature, PTF $\chi^2$  : PTF selected by $\chi^2$ , $\Delta KL$ PTF: PTF selected via $\Delta KL$}
  \label{tab:addlabel}%
   
\end{table}%
\section{Spy Induction}
We recall from section 1 that our problem suffers with limited training data per author as sockpuppets only use an alias few times. To improve verification, we need a way to learn from more instances. Also from section 4, we know that precision drops with increase in diversity of $\neg a$. This can be addressed by leveraging the unlabeled test set to improve the $\neg a$ set in training under transduction. \par
Figure 2 provides an overview of the scheme. For a given training set and a test set for $AV_a$, spy induction has three main steps. First is spy selection where some carefully selected positive samples are sent to the unlabeled test set. The second step is to find certain Nearest and Farthest Neighbors (abbreviated NN, FN henceforth) of the positive spy samples in the unlabeled test set. As the instances retain their original identity, a good distance metric should be able to retrieve potentially hidden positive (using common NN across different positive spies) and negative (using common FN across different positive spies) samples in the unlabeled test set. These newly retrieved samples from unlabeled test set are used to grow the training set. The previous step can have some label errors in NN and FN as they may not be true positive ($a$) and negative ($\neg a$) samples, which can be harmful in training. These are shown in Figure 2(B) by $\alpha_-$ and $\beta_+$ samples. To reduce such potential errors, a third step of label verification is employed where the labels of the newly retrieved samples from unlabeled test set are verified using agreement of classifiers on orthogonal feature spaces. with this step, we benefit from the extended training data without suffering from the possible issue of error propagation. Lastly, the verifier undergoes improved training with additional samples and optimizes the F-Score on the training set.\par 
\begin{figure}[t!]
\includegraphics[scale=.95]{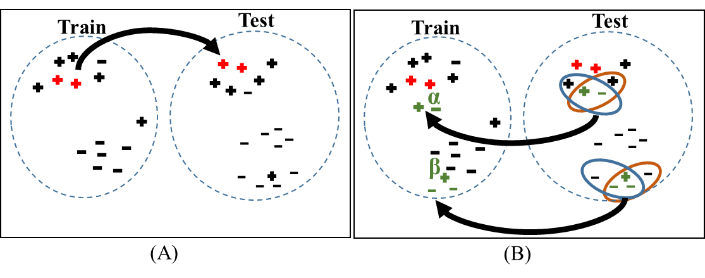}
\caption{Spy Induction : (A)Spies (Red plus signs) selected based on positive class centrality being put to the unlabeled test set. (B) Common nearest and farthest neighbors (Green plus and minus signs) across different spies’ neighborhood shown by oval boundaries found in unlabeled test set being put back in the training set.}
\end{figure}
\subsection{Spy Selection}
This first step involves sending highly representative spies that can retrieve new samples to improve training. For a given verification problem, $AV_a$, let $D=D.Train\cup D.Test$ denotes the whole data. Although any positive instance in $D.Train$ can be a spy sample, only few of them might satisfy the representativeness constraint. Hence, we select the spies as those positive samples that have maximum similarity with other positive instances. In other words, the selection respects class based centrality and employs minimum overall pairwise distance (OPD) as its selection criterion:
\begin{equation}
OPD(s)=argmin_{s\in P} (\sum_{x\in P}d(s,x))
\end{equation}\par
where $P$ is the positive class of training set, $s$ denotes a potential spy sample and $d(\cdot)$ is distance function. Our spy set, $S=\{s\}$ consists of different spies that have the least pairwise distance to all other positive samples. We also consider different sizes of the spy set $|S|=n_S$ and experiment with different values of $n_S\in N_S=\{1,3,5,7\}$. The method $SelectSpy(\cdot)$ (line 4, Algorithm 1) implements this step.
\subsection{New Instance Retrieval via Nearest and Farthest Neighbors}
After the selected spies are put into the unlabeled test set, the goal is to find potential positive and negative samples. Intuitively, one would expect that the closest data points to positive spy samples belong to the positive class while those that are farthest are likely negative samples. For each spy, $s\in S$, we consider $n_Q$  nearest neighbors forming the likely positive set $Q_s$ and $n_R$ farthest neighbors forming the likely negative set $R_s$ specific to $s$. Then, we find the common neighbors across multiple spies to get confidence on the likely positive or negative samples which yields us the final set of potentially $Q$ positive and $R$ negative samples,
\begin{equation}
Q=\cap_{s\in S}\, Q_s ;\quad R=\cap_{s\in S}\, R_s 
\end{equation}\par
This is implemented by the methods $ObtainNN(\cdot),ObtainFN(\cdot)$ (lines 5, 6, Algorithm 1). In most cases, we did not find the common neighbors $Q,R$ to be empty, but if it is null, it implies no reliable samples were found. Further, like $n_S$ (in section 6.1), we try different values for $|Q_s |=n_Q; n_Q \in N_Q=\{1,3\}$ and $|R_s |=n_R; n_R \in N_R=\{5,10,25,40,50,60\}$. These values were set based on pilot experiments.
The above scheme of new sample retrieval works with any distance metric. \par We consider two distance metrics on the feature space L+$\Delta KL$ PTF to compute all pairwise distances in the methods $SelectSpy(\cdot)$, $ObtainNN(\cdot)$ and $ObtainFN(\cdot)$ (line 4-6, Algorithm 1): (1) Euclidean, (2) Distance metric learned from data. Specifically, we use the large margin method in \cite{weinberger2005distance} which learns a Mahalanobis distance metric $d_M(\cdot)$ that optimizes kNN classification in the training data using $d_M$. The goal is to learn $d_M(\cdot)$ such that the k-nearest neighbors (based on $d_M(\cdot)$) of each sample have the same class label as itself while different class samples are separated by a large margin.
\begin{figure}[t!]
\hrule
{\strut\footnotesize  Algorithm 1: Spy induction} 
\hrule
\footnotesize 
\begin{flushleft}
$ SpyInduction(D,N_S,N_Q,N_R)$\\
$ 1:P\leftarrow \{x \in D.Train, x.label>0\}//positive\,class$\\
$ 2 :I\leftarrow \{(n_S,n_Q,n_R)|n_S\in $ $ N_S,n_Q\in N_Q,n_R\in N_R\} $\\
$ 3 :for\ each (i=(n_S,n_Q,n_R )\in$ $ I)$\\
 $ 4 :\quad	S\leftarrow SelectSpy(P,n_S)$\\
 $ 5 :\quad    Q\leftarrow ObtainNN(D.Test,S,n_Q)$\\
 $ 6 :\quad    R\leftarrow ObtainFN(D.Test,S,n_R)$\\
 $ 7 :\quad    (Q^v,R^v)\leftarrow CoLabelingVerification(Q,R,D.Train)$\\
 $ 8 :\quad     F1(i)\leftarrow CVImprovedTraining(D.Train,Q^v,R^v)$\\
 $ 9 :endfor$\\
$ 10:(n_S,n_Q,n_R)^*\leftarrow {argmax}\ _{i\in I}{(F1(i))}$\\
$ 11:AV\leftarrow Classifier(D.Train,D.Test,(n_S,n_Q,n_R)^*)$\\
\end{flushleft}
\normalsize
\hrule
\caption{Spy induction}
\label{fig:algorithm1}
\end{figure}
\subsection{Label Verification via Co-Labeling}
As it is not guaranteed that the distances between samples can capture the notion of authorship, the previous step can have errors, i.e., there may be some positive samples in $R$ and negative samples in $Q$. To solve this, we apply co-labeling \cite{xu2015co} for label verification. In co-labeling, multiple views are considered for the data and classifiers are built on each view. Majority voting based on classifier agreement is used to predict labels of unlabeled instances. In our case, we consider $D.Train$ to train an SVM on five feature spaces (views): i) unigam, ii) unigram+bigram, iii) PTF, iv) POS , v) $\Delta KL$ PTF+unigram+bigram as five different label verification classifiers. Then, the labels of samples in $Q$ and $R$ are verified based on agreements of majority on classifier prediction. Samples having label discrepancies are discarded to yield the verified retrieved samples, $(Q^v,R^v)$ (line 7, Algorithm 1). The rationale here is that it is less probable for majority of classifiers (each trained on a different view) to make the same mistake in predicting the label of a data point than a single classifier.
\subsection{Improved Training}
The retrieved and verified samples from the previous steps are put back into the training set. However, the key lies in estimating the right balance between the amount of spies sent, and the size of the neighborhood considered for retrieving potentially positive or negative samples, which are governed by the parameters $n_S,n_Q,n_R$. To find the optimal parameters, we try different values of the parameter triple, $i=(n_S,n_Q,n_R )\in I$ (lines 2, 3 Algorithm 1) and record the F-Score of 5-fold CV on $D.Train\cup Q^v\cup R^v$ as $F1(i)$ (line 8, Algorithm 1). This step is carried out by the method $CVImprovedTraining(.)$. Finally, the parameters that yield the highest $F1$ in training are chosen (line 10, Algorithm 1) to yield the output spy induced verifier (line 11, Algorithm 1).
\section{Experimental Evaluation}
This section evaluates the proposed spy method. We keep all experiment settings same as in section $5$ (i.e., use out-of-training with varying author diversity $\delta$). We fix our feature space to L+$\Delta KL$ PTF as it performed best (see Table 3). As mentioned earlier, we report average verification performance across all authors. Below we detail baselines, followed by results and sensitivity analysis.
\subsection{Baselines and Systems}
We consider the following systems:\\
\textbf{MBSP} runs the Memory-based shallow parsing approach \cite{luyckx2008authorship} to authorship verification that is tailored for short text and limited training data.\\
\textbf{Base} runs classification without spy induction and dovetails with Table 3 (last row) for each $\delta$.\\
\textbf{TSVM} uses the transductive learner of SVMLight  \cite{joachims1999transductive} and aims to leverage the unlabeled (test) set by classifying a fraction of unlabeled samples to the positive class and optimizes the precision/recall breakeven point.\\
\textbf{Spy (Eu.) \& Spy (LM)} are spy induction systems without co-labeling but use Euclidean (Eu.) and learned distance metric (LM) to compute neighbors.\\
\textbf{Spy (EuC) \& Spy (LMC)} are extensions of previous models that consider label verification via co-labeling approach.
\begin{table}[t!]

  \centering
  
  \scalebox{1}{
    \begin{tabular}{lrrrrrrrrrc}
    
    \toprule
    \multicolumn{10}{c}{ \bf{$\delta$=25\%}}                                                   &  \\
    
          & \multicolumn{3}{c}{\bf{SVM}} & \multicolumn{3}{c}{\bf{LR}} & \multicolumn{3}{c}{\bf{kNN}} & \bf{AVG} \\
    Model & \multicolumn{1}{c}{P} & \multicolumn{1}{c}{R} & \multicolumn{1}{c}{F1} & \multicolumn{1}{c}{P} & \multicolumn{1}{c}{R} & \multicolumn{1}{c}{F1} & \multicolumn{1}{c}{P} & \multicolumn{1}{c}{R} & \multicolumn{1}{c}{F1} & F1 \\
     \midrule
      MBSP  & 22.9  & 84.1  & 32.0  & 22.1  & 82.1  & 31.1  & 20.7  & 77.5  & 23.5  & 28.9 \\
    Base  & 25.6  & 79.2  & 36.3  & 21.7  & 80.2  & 32.1  & 22.3  & 81.5  & 27.8  & 32.1 \\
    TSVM  & 30.6  & 43.9  & 34.3  & \multicolumn{1}{c}{-} & \multicolumn{1}{c}{-} & \multicolumn{1}{c}{-} & \multicolumn{1}{c}{-} & \multicolumn{1}{c}{-} & \multicolumn{1}{c}{-} & 34.3 \\
    Spy(Eu.) & 39.1  & 51.2  & 40.6  & 51.6  & \underline{42.3}  & 43.7  & 43.3  & \underline{52.5}  & 39.4  & 41.2 \\
    Spy(LM) & 42.2  & 49.3  & 42.0    & 44.4  & \underline{49.6}  & 43.9  & 34.9  & \underline{62.8}  & 33.7  & 39.9 \\
    Spy(EuC) & 42.0    & \underline{57.8}  & 42.7  & 51.2  & \underline{61.3}  & \colorbox{gray!50}{52.5}  & 41.5  & \underline{57.9}  & 38.4  & \textbf{44.5} \\
    Spy(LMC) & 38.1  & \underline{60.5}  & 40.6  & 42.9  & \underline{64.9}  & 47.1  & 35.3  & \underline{68.1}  & 36.0    & 41.2 \\
    \multicolumn{10}{c}{(A)}                                                      &  \\
    \midrule
    \multicolumn{10}{c}{ \bf{$\delta$=50\%}}                                                   &  \\
          &\multicolumn{3}{c}{\bf{SVM}} & \multicolumn{3}{c}{\bf{LR}} & \multicolumn{3}{c}{\bf{kNN}} & \bf{AVG} \\
    Model & \multicolumn{1}{c}{P} & \multicolumn{1}{c}{R} & \multicolumn{1}{c}{F1} & \multicolumn{1}{c}{P} & \multicolumn{1}{c}{R} & \multicolumn{1}{c}{F1} & \multicolumn{1}{c}{P} & \multicolumn{1}{c}{R} & \multicolumn{1}{c}{F1} & F1 \\
     \midrule
     MBSP  & 31.9  & 85.3  & 42.1  & 25.0  & 81.6  & 34.6  & 21.1  & 84.4  & 28.7  & 35.1 \\
    Base  & 33.7  & 75.9  & 42.8  & 31.1  & 79.4  & 41.9  & 26.9  & 79.5  & 30.3  & 38.3 \\
    TSVM  & 20.2  & 83.6  & 31.1  & \multicolumn{1}{c}{-} & \multicolumn{1}{c}{-} & \multicolumn{1}{c}{-} & \multicolumn{1}{c}{-} & \multicolumn{1}{c}{-} & \multicolumn{1}{c}{-} & 31.1 \\
    Spy(Eu.) & 39.1  & 71.2  & 45.9  & 38.1  & 67.9  & 46.2  & 45.1  & \underline{58.7}  & 41.7  & 44.6 \\
    Spy(LM) & 40.1  & 68.5  & 45.9  & 44.7  & 55.5  & 45.6  & 42.7  & \underline{66.2}  & 40.2  & 43.9 \\
    Spy(EuC) & 62.3  & \underline{52.0}    & 52.3  & 62.5  & \underline{64.6}  & \colorbox{gray!50}{61.0}    & 46.6  & \underline{62.3}  & 43.2  & \textbf{52.2} \\
    Spy(LMC) & 46.8  & \underline{60.9}  & 48.0    & 51.5  & \underline{67.4}  & 53.7  & 40.7  & \underline{67.6}  & 39.2  & 47.0 \\
    \multicolumn{10}{c}{(B)}                                                      &  \\
    \midrule
    \multicolumn{10}{c}{\bf{$\delta$=75\%}}                                                   &  \\
          & \multicolumn{3}{c}{\bf{SVM}} & \multicolumn{3}{c}{\bf{LR}} & \multicolumn{3}{c}{\bf{kNN}} & \bf{AVG} \\
    Model & \multicolumn{1}{c}{P} & \multicolumn{1}{c}{R} & \multicolumn{1}{c}{F1} & \multicolumn{1}{c}{P} & \multicolumn{1}{c}{R} & \multicolumn{1}{c}{F1} & \multicolumn{1}{c}{P} & \multicolumn{1}{c}{R} & \multicolumn{1}{c}{F1} & F1 \\
     \midrule
     MBSP  & 49.9  & 80.4  & 57.2  & 53.9  & 81.9  & 59.1  & 33.8  & 82.2  & 38.6  & 51.6 \\
    Base  & 50.5  & 71.9  & 56.2  & 46.3  & 79.4  & 54.9  & 42.2  & 80.8  & 46.1  & 52.4 \\
    TSVM  & 34.4  & 80.4  & 45.8  & \multicolumn{1}{c}{-} & \multicolumn{1}{c}{-} & \multicolumn{1}{c}{-} & \multicolumn{1}{c}{-} & \multicolumn{1}{c}{-} & \multicolumn{1}{c}{-} & 45.8 \\
    Spy(Eu.) & 55.6  & 70.8  & 58.2  & 50.9  & 77.5  & 57.9  & 57.7  & 57.6  & 50.7  & 55.6 \\
    Spy(LM) & 53.1  & 62.8  & 54.3  & 51.1  & 69.6  & 56.1  & 51.8  & 57.7  & 45.8  & 52.1 \\
    Spy(EuC) & 71.9  & \underline{59.1}  & 62.4  & 68.9  & 75.6  & \colorbox{gray!50}{70.2}  & 63.6  & \underline{59.0}    & 54.7  & \textbf{62.4} \\
    Spy(LMC) & 55.6  & \underline{72.3}  & 58.4  & 60.8  & 68.5  & 61.4  & 53.4  & \underline{60.8}  & 48.7  & 56.2 \\
    \multicolumn{10}{c}{(C)}                                                      &  \\
    \bottomrule
    \end{tabular}%
   
    }
 \caption{P: Precision, R: Recall, F1: F-Score results for spy induction under out-of-training with different values of $\delta$ for three classifiers. AVG reports the average F1 across three classification models. Feature Set: L+$\Delta KL$ PTF. Gains in AVG F1 using spy (EuC) and (LMC) over baselines are significant at p$<$0.001 using a t-test}
\end{table}%
\subsection{Results}
Table 4 reports the results. We note the following:
\begin{itemize}[leftmargin=*]

\item Except for two cases (F1 of SVM and kNN for Spy(LM) with $\delta=75\%$), almost all spy models are able to achieve significantly higher F1 than base (without spy induction) and TSVM for all classifiers SVM, LR, kNN and across all diversity values $\delta$. MBSP performs similarly as Base showing memory based learning does not yield a significant advantage in sockpuppet verification. TSVM is not doing well on F1 but improves recall. One reason could be that due to class imbalance, TSVM has some bias in classifying unlabeled examples to positive class that improves recall but suffers in precision.
\item The AVG F1 column shows that on average, across three classifiers spy induction yields at least 4\% gain or more. The gains in AVG F1 are pronounced for $\delta =25\%$ with gains upto 12\% with spy (EuC). For $\delta =75\%$, we find gains of about 10\% in F1 with spy (EuC). Note that we employ out-of-training setting with varying author diversity ($\delta$) so the test set is imbalanced (i.e., the random baseline is no longer 50\%). Across all classifiers, the relative gains in F1 for spy methods over base reduce with increase in author diversity $\delta$ which is due (a) better $\neg a$ samples in training that raise the base result and (b) test set size and variety reduction limiting spy induction. Nonetheless, we note that for $\delta=25\%$ (harder case of verification), spy induction does well across all classifiers.
\item Anchoring on one distance metric (Eu./LM), we find that spy induction with co-labeling does markedly better than spy induction without co-labeling across all $\delta$ in AVG F1 across three classifiers. This shows label verification using co-labeling is helpful in filtering label noise and an essential component in spy induction.
\item Between Euclidean and distance metric learned via large-margin (LM), Euclidean does better than LM in AVG F1 for both spy induction with and without co-labeling. However, using the LM metric yields higher recall than Euclidean in certain cases (underlined) which shows LM metric can yield gains in F1 beyond base with relatively lesser drops in recall which is again useful.
\end{itemize}\par
In summary, we can see that spy induction works in improving the F1 across different classifiers and author diversity and distance metrics. Overall, the scheme LR+Spy (EuC) does best across each $\delta$ (highlighted in gray) and is used for subsequent experiments to compare against Base.	
\subsection{Spy Parameter Sensitivity Analysis}
To analyze the sensitivity of the parameters, we plot the range of precision, recall and F1 values as spy induction learns the optimal values in training. We focus on the variation for $\delta =25\%,75\%$ capturing both extremes of diversity. Figure 3 shows the 
performance curves for different spy parameter triples $(n_S,n_Q,n_R)$ sorted in the increasing order of F1. We find that for both $\delta =25\%,75\%,$ the spy induction steadily improves precision with the increase in likely $\neg a$ samples $(n_R$). Although the recall drops more and has more fluctuations for the harder case of $\delta =25\%$, it stabilizes early for $\delta =75\%$ with much lesser drop in recall. This shows that the spy induction scheme is robust in optimizing F1 with only a few (5-7) spy samples $(n_S)$ sent to unlabeled test set.
\begin{table}[t!]

  \centering
  
  \scalebox{1}{
    \begin{tabular}{lcccccc}
    \toprule
    \multirow{2}[4]{*}{} & \multicolumn{2}{c}{$\delta$ =25\%} & \multicolumn{2}{c}{ $\delta$=50\%} & \multicolumn{2}{c}{$\delta$ =75\%} \\
    \midrule
          & Base  & Spy   & Base  & Spy   & Base  & Spy \\
    Test Domain & F1    & F1    & F1    & F1    & F1    & F1 \\
    Hotel & 30.3  & 36.4  & 40.0    & 47.0    & 50.3  & 52.6 \\
    Product & 29.5  & 36.6  & 34.0    & 40.8  & 51.5  & 53.5 \\
    Restaurant & 30.1  & 41.1  & 41.7  & 51.5  & 55.2  & 59.3 \\
    \bottomrule
    \end{tabular}%
  \label{tab:addlabel}%
  }
  \caption{ Cross domains results of LR + Spy (EuC). Gains in F1 using spy induction over base are significant at p$<$0.01 for all test domains and each $\delta$ using a t-test.}
\end{table}%
\begin{figure}
\includegraphics[scale=.275]{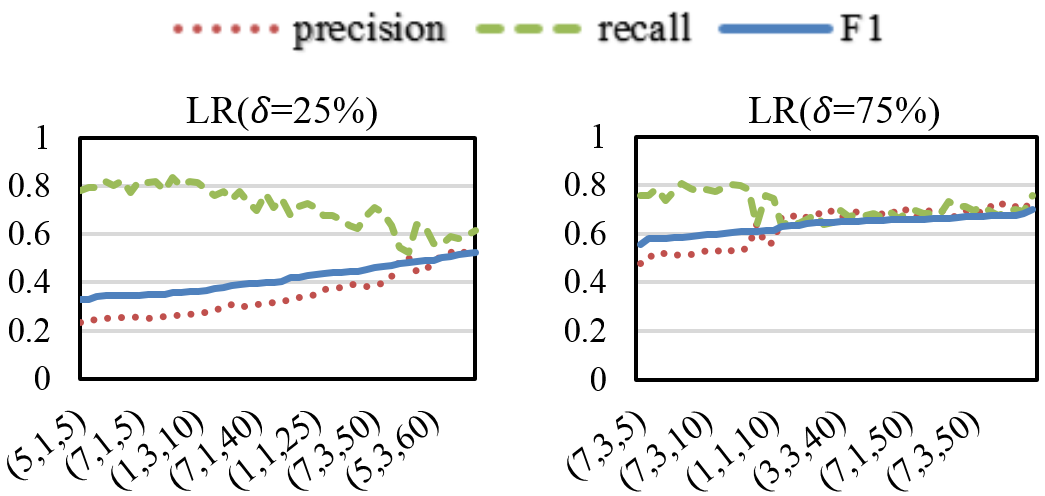}
\caption{Spy Parameter Sensitivity. Variation of precision, recall and F1 across different parameter triples $(n_S,n_Q,n_R)$.}
\end{figure}
\subsection{Domain Adaptation}
We now test the effective of spy induction under domain transfer. As mentioned in section 3, we obtained reviews of Turkers for hotel, restaurant and product domains. Keeping all other settings same as in Table 4, Table 5 reports results for cross domain performance by training the verifiers ($AV_a$) using two domains and testing on the third domain. We compare sockpupet verification using LR+Spy (EuC) vs. base (LR without spy induction). We report the F1 scores as the trends of precision and recall for cross domain were similar to the trends in Table 4. The F1 of base in cross domain (Table 5, Hotel row) is lower than corresponding LR results with base (Table 4) for all $\delta$ showing cross domain verification is harder. Nonetheless, spy induction is able to render statistically significant gains in F1 for all $\delta$ (see Table 5).
\begin{table}[t!]
  
  \centering

    \begin{tabular}{lcccccc}
    \toprule
    \multirow{2}[4]{*}{} & \multicolumn{2}{c}{$\delta$ =25\%} & \multicolumn{2}{c}{ $\delta$=50\%} & \multicolumn{2}{c}{$\delta$ =75\%} \\
    \midrule
          & Base  & Spy   & Base  & Spy   & Base  & Spy \\
    Classifier & F1    & F1    & F1    & F1    & F1    & F1 \\
    SVM & 50.7  & 57.6  & 59.6    & 62.9    & 68.0  & 70.3 \\
    LR & 40.4  & 42.4  & 49.8    & 51.1  & 60.0  & 61.5 \\
    kNN & 23.9  & 29.5  & 32.0  & 37.1  & 43.0  & 51.1 \\
    \bottomrule
    \end{tabular}%
  \label{tab:addlabel}%
 
  \caption{ Performance gains of Spy (EuC) in F1 over Base on Wikipedia Sockpuppet Dataset. Gains are significant (p$<$0.01) except for LR δ =50\%, 75\% }

\end{table}%

  \subsection{Performance on Wikipedia Sockpuppet (WikiSock) Dataset}
In \cite{SOLORIO14.1007}, a corpus of Wikipedia sockpuppet authors was produced. It contains 305 authors with an average of 180 documents per author and 90 words per document which we use as another benchmark for evaluating our method.\\ It is important to note that the base results reported in \cite{SOLORIO14.1007} are not directly comparable to this experiment (Table 6). This is because \cite{SOLORIO14.1007} used all 623 cases that were found as candidates but we focus on only 305 of them which were actually confirmed sockpuppets by Wikipedia administrators. Next, we perform experiments under realistic out-of-training setting and varying the author diversity (as in Table 4) which is different from \cite{SOLORIO14.1007}. This explains the rather lower F1 as reported in \cite{SOLORIO14.1007} for Base. We focus on F1 performance of spy (EuC) versus base (without spy) as the precision and recall trends were same as in Table 4. Compared to Table 4 base results, base does better for SVM and LR on WikiSock dataset that  hints the data to be slightly easier. The relative gains of spy over base although are a bit lower than those in Table 4, spy induction consistently outperforms base.
\section{Conclusion}
This work performed an in-depth analysis of deceptive sockpuppet detection. We first showed that the problem is different from traditional authorship attribution or verification and gets more difficult with the increase in author diversity. Next, a feature selection scheme based on KL-Divergence of stylistic language models was explored that yielded improvements in verification beyond baseline features. Finally, a transduction scheme, spy induction, was proposed to leverage the unlabeled test set. A comprehensive set of experiments showed that the proposed approach is robust across both (1) different classifiers, (2) cross domain knowledge transfer and significantly outperforms baselines. Further, this work produced a ground truth corpus of deceptive sockpuppets across three domains.
\section*{Acknowledgments}
This work is supported in part by NSF 1527364. We also thank anonymous reviewers for their helpful feedbacks.
\bibliographystyle{splncs}
\bibliography{paper}
\end{document}